\documentclass[conference]{IEEEtran}
\IEEEoverridecommandlockouts

\usepackage{titlesec}
\titleformat{\subsubsection}[runin] 
{\normalfont\itshape} 
 {\hspace{1em}\arabic{subsubsection})} 
{0.3em} 
{} 
\titlespacing*{\subsubsection}{0pt}{0.5em}{0.3em}  

\usepackage{soul}   
\usepackage{xcolor} 
\usepackage{threeparttable}
\usepackage{float}
\usepackage[ruled,vlined]{algorithm2e}
\usepackage{multirow}
\usepackage{array}
\usepackage{booktabs}
\usepackage{cite}
\usepackage{amsmath,amssymb,amsfonts}
\usepackage{algorithmic}
\usepackage{graphicx}
\usepackage{textcomp}
\usepackage{siunitx}
\usepackage{tabularx}
\usepackage{ifthen}
\usepackage{url}

\graphicspath{{./figures/}{./figures/results}} 

\usepackage{cleveref} 
\crefname{section}{§\!\!}{§§\!\!}
\Crefname{section}{Section}{Sections}
\crefname{figure}{Fig.}{Figs.}
\Crefname{figure}{Figure}{Figures}
\crefname{table}{Table}{Tables}
\Crefname{table}{Table}{Tables}
\usepackage[inline,shortlabels]{enumitem} 

\def\BibTeX{{\rm B\kern-.05em{\sc i\kern-.025em b}\kern-.08em
    T\kern-.1667em\lower.7ex\hbox{E}\kern-.125emX}}

\def\eg{e.g., }
\def\ie{i.e., }
\def\vs{vs.\ } 
\def\cf{cf.\ } 

\def\trackchanges{0} 
\ifthenelse{\equal{\trackchanges}{1}}{
    \newcommand{\newtext}[1]{\textcolor{blue}{#1}}
    \newcommand{\oldtext}[1]{\textcolor{purple}{\st{#1}} }
}{
    \newcommand{\newtext}[1]{{#1}}
    \newcommand{\oldtext}[1]{}
} 
\newcommand{\replacetext}[2]{\oldtext{#1}\newtext{#2}}

\begin{document}



\title{
Generalization of 
\replacetext{GNN}{Graph Neural Network}
Models \\for Distribution Grid Fault Detection 
}

\author{
\IEEEauthorblockN{
    Burak~Karabulut$^{\dagger,\star}$ \quad \quad \quad
    Carlo~Manna$^{\star}$ \quad \quad \quad
    Chris~Develder$^\dagger$
}
\vspace{5pt}
\IEEEauthorblockA{
\begin{tabular}{cc}
$^\dagger$\textit{Dept.\ of Information Technology, IDLab}
  &\quad $^\star$\textit{Water and Energy Transition Unit} \\
\textit{Ghent University -- imec}, Ghent, Belgium 
  &\quad \textit{VITO}, Mol, Belgium \\
\{burak.karabulut,\,chris.develder\}@ugent.be
    &\quad \{burak.karabulut,\,carlo.manna\}@vito.be
\end{tabular}}
}

\maketitle

\begin{abstract}
Fault detection in power distribution grids is critical for ensuring system reliability and preventing costly outages. Moreover, fault detection methodologies should remain robust to evolving grid topologies caused by factors such as reconfigurations, equipment failures, and Distributed Energy Resource (DER) integration.
Current data-driven state-of-the-art methods 
use Recurrent Neural Networks (RNNs) for temporal modeling and Graph Neural Networks (GNNs) for spatial learning, in an RNN+GNN pipeline setting (RGNN in short). Specifically, for power system fault diagnosis, Graph Convolutional Networks (GCNs) have been adopted. Yet, various more advanced GNN architectures have been proposed and adopted in domains outside of power systems. 
In this paper, we set out to systematically and consistently benchmark various GNN architectures in an RNN+GNN pipeline model. Specifically, to the best of our knowledge, we are the first to 
\begin{enumerate*}[(i)]
    \item propose to use GraphSAGE and Graph Attention (GAT, GATv2) in an RGNN for fault diagnosis, and
    \item provide a comprehensive benchmark against earlier proposed RGNN solutions (RGCN) \replacetext{and}{as well as} pure RNN models (\newtext{especially} \replacetext{GRU}{Gated Recurrent Unit (GRU)}), particularly
    \item exploring their generalization potential for deployment in different settings than those used for training them.
\end{enumerate*}
Our experimental results on the IEEE 123-node distribution network show that RGATv2 has superior generalization capabilities, maintaining high performance \newtext{with an F1-score reduction of $\sim$12\%} across different topology settings. \newtext{In contrast, pure RNN models largely fail, experiencing an F1-score reduction of up to $\sim$60\%, while other RGNN variants also exhibit significant performance degradation, \ie up to $\sim$25\% lower F1-scores.}
\end{abstract}


\section{Introduction}
\label{sec:intro}
Fault diagnosis in power distribution grids is essential for maintaining grid reliability and preventing costly outages~\cite{chen2020fault}. In particular, faults, typically involving short circuits caused by various factors such as weather, equipment failures, or insulation breakdowns~\cite{teixeira2021fault}, must be detected and localized quickly, often within milliseconds, to avoid further damage in the grid and minimize downtime for users. Such fault detection today has become way more challenging, due to increased
\begin{enumerate*}[(1)]
   \item deployment of DERs, and 
   \item electrification (e.g., EV chargers)
\end{enumerate*}.
First of all, they change the conventional fault propagation patterns; \eg faults may propagate from multiple directions within the grid as the power flow becomes bidirectional\cite{karimi2021spatiotemporal}. Second, inherent variability and intermittency~\cite{ren2021quantitative} of renewable sources (\eg variations in generation due to weather conditions) can lead to voltage fluctuations and load imbalances~\cite{adetokun2023impact}, and thus amplify the need for adaptable fault diagnosis strategies.
 

Besides the increased complexity of (distribution) grids in terms of distributed generation and new loads, their topological configuration also changes. Indeed, distribution grids operate under diverse and evolving topologies due to switching operations, network reconfigurations, and equipment failures~\cite{teixeira2021fault}. Also, new \replacetext{monitoring}{measurement} devices \newtext{such as Phasor Measurement Units (PMUs) and smart meters,} may be introduced, while existing ones can malfunction or be retired, requiring continuous updates to \replacetext{monitoring}{measurement} configurations~\cite{ojha2018optimization}. Thus, effective fault diagnosis methods should not only be accurate and robust, but also able to adapt to evolving grid structures and diverse configurations. Such fault diagnosis in power systems comprises three tasks: \emph{fault detection} determines whether a fault is present, \emph{fault classification} identifies the fault type or affected phase, and \emph{fault localization} pinpoints the fault’s location in the network~\cite{nguyen2023spatial}. Existing methods for fault diagnosis in the power system literature~\cite{chen2016fault, ganivada2022fault, ju2021fault}, can be broadly classified as model-based methods and data-driven methods, which we discuss in turn in the next paragraphs.

\textit{Model-based methods} can be categorized into analytical methods, protection-based methods, and signal-based methods. Analytical methods use mathematical models, such as impedance-based approaches~\cite{chen2020fault, wang2024enhanced, tavoosi2022hybrid}, matrix-based approaches~\cite{mansourlakouraj2021application}, and voltage-sag-based approaches~\cite{chen2020fault, turizo2022voltage}, to estimate fault location and infer fault type using voltage and current measurements. Protection-based methods, including overcurrent relays, differential protection, and directional relays, operate on predefined thresholds and relay-based schemes to detect and classify faults based on deviations from normal operating conditions. Signal-based methods focus on analyzing  high-frequency transient signals to detect and locate faults, such as traveling wave analysis and wavelet-based techniques. Traveling wave methods are particularly effective for fault detection and localization, but require high-frequency sensors, which increase deployment costs and may not be feasible in distribution systems~\cite{mansourlakouraj2021application, liu2023novel}. Model-based methods typically use predefined thresholds, heuristic rules, and analytical models to detect, classify, and locate faults~\cite{wang2024enhanced, lakouraj2021graph, chen2020fault}. Although model-based methods are effective in conventional grid settings, they face challenges (e.g., false alarms, reduced performance in grids with distributed generation) in modern distribution networks due to their reliance on static assumptions about grid topology, operating conditions, and fault characteristics. 

To address such challenges, early \textit{data-driven methods} have been increasingly adopted to overcome the limitations of traditional model-based approaches, e.g.,~\cite{livani2013faulty,livani2013FaultClass, hosseini2018ami, fahim2021deep}. Early artificial intelligence (AI) approaches, such as decision trees, k-nearest neighbors (kNN), and support vector machines (SVMs), have been used to classify faults based on manually extracted features~\cite{chen2016fault}.
More recent models based on deep learning, such as 
Convolutional Neural Networks (CNNs) and Recurrent Neural Networks (RNNs), have shown promise by directly learning patterns from raw or minimally processed data.
CNNs have been particularly effective in analyzing spatial features, while RNNs (e.g., Long Short-Term Memory (LSTM) or GRU models) leverage temporal dependencies in sequential data~\cite{nguyen2023spatial}.
However, traditional AI models are designed primarily for fairly rigorously structured data (e.g., matrices of image pixels, time series at a fixed measurement frequency). 
Power systems, in contrast, are 
networks of interconnected nodes (buses) and edges (power lines), where relationships between data points are defined by non-uniform electrical connectivity rather than fixed spatial arrangements such as physical distance.
Thus, methods such as CNNs struggle to fully exploit the underlying structure of the grid, particularly as new measurement configurations emerge due to the integration of DERs~\cite{lakouraj2021graph, liao2022review}.

Graph learning approaches, as GNNs pioneered by Gori et al.~\cite{gori2005new, scarselli2009graph}, have been increasingly adopted to overcome aforementioned limitations of traditional AI models.
GNNs are specifically designed to process data structured as graphs: power grid components such as buses and their interconnecting distribution power lines can naturally be represented as graph nodes and edges. 
Using message-passing algorithms, GNNs are able to capture dependencies between nodes and edges~\cite{liao2022review}.
MansourLakouraj et al.~\cite{mansourlakouraj2021application} used spectral Graph Convolutional Networks (GCNs) for fault localization by leveraging the graph structure of the power system to model spatial dependencies. Spectral GCNs operate on the graph’s Laplacian matrix, thus allowing for the incorporation of the global graph structure. Spectral methods require recalculating the entire graph at each step, making them computationally expensive and less efficient for the tasks that focus on local graph structures rather than global ones~\cite{liao2022review}. 

On the other hand, unlike spectral methods, spatial GCNs are more efficient as they operate directly on the graph's node and edge features, aggregating information from neighboring nodes to capture local dependencies and avoiding the redundant recalculation of the entire graph. Hence, for tasks such as fault localization, spatial GCNs are often preferred. For example,~\cite{lakouraj2021graph} and~\cite{ma2024method} used spatial GCNs, specifically from the framework proposed by Kipf and Welling~\cite{kipf2016semi}, for the fault localization task. In~\cite{teixeira2021fault}, the authors propose a 
\replacetext{gated graph neural network}{Gated Graph Neural Network}
(GGNN) that leverages GNNs, while also using gated mechanisms to update the hidden states of the nodes over time.
However, both of these approaches do not account for the time-series nature of fault events, as the proposed methodology models only spatial relationships.
In contrast, Bang et al.~\cite{nguyen2022one} 
experimented with a 1D-Convolutional GNN, where they use 1D-CNN to capture temporal dependencies, and then a GCN~\cite{kipf2016semi} to extract spatial correlations for fault diagnosis tasks.
Later, in~\cite{nguyen2023spatial}, 
they enhanced this hybrid approach by incorporating Long Short-Term Memory (LSTM) networks. Their approach effectively captures both spatial and temporal dependencies.  

However, 
the adaptability of GCNs [26] to evolving grid topologies 
is limited due to the reliance of the model on a fixed adjacency matrix, 
preventing it from generalizing to unseen nodes when the graph structure changes. 
In contrast, the work proposed in~\cite{wang2024enhanced} incorporates Graph Attention Networks (GATs), which use an attention mechanism to learn node importance based on the local graph structure.
This ability to focus on relevant nodes in the graph, rather than relying on a fixed adjacency matrix to calculate node representations, makes GATs more adaptable to the changes in graph topology.
Ngo et al.~\cite{ngo2024deep} extend GAT models by incorporating 1D-CNN to extract temporal information to improve fault diagnosis accuracy by using both voltage and branch current inputs.
However, the robustness of their model under evolving grid conditions remains untested.



In conclusion, GNN-based models show great promise for distribution grid fault diagnosis, yet
\begin{enumerate*}[(i)]
    \item there has not been a rigorous performance comparison of the various recent GNN architectures for fault diagnosis, and
    \item their conceptual potential to deal with variations/extensions of the grid in terms of 
    \replacetext{monitors}{measurement devices} and/or topological changes has not been studied.
\end{enumerate*}
The current paper forms a pilot study to try and address these issues, in particular by
\begin{enumerate}[(1)]
    \item Proposing RNN+GNN architectures previously unexplored for power system fault detection \ie the RGSAGE and RGAT models; 
    \item Conducting a systematic quantitative benchmarking of the proposed models against the state-of-the-art RGCN model as well as non-GNN models (based on GRUs); and
    \item Evaluating \newtext{the} generalization capabilities of the various GNN-based solutions for deployments with varying number of \replacetext{monitors}{PMUs}, on a realistic, sizeable distribution grid configuration (the IEEE 123-node network).
\end{enumerate}
We also discuss next steps by which we will extend the current pilot to a broader scope of topological variations and fault diagnosis applications. 



\section{Methodology}
\label{sec:method}
The limitations identified from the literature review above highlight the need for a structured evaluation of GNN architectures and their ability to handle evolving grid topologies.
Next, we discuss the RNN+GNN pipeline components and particularly the GNN models in more detail.

\subsection{Temporal Feature Extraction -- Recurrent Neural Networks}
Fault data in distribution grids inherently constitute time series data. 
To capture these temporal dependencies, recurrent neural networks (RNNs) are widely used~\cite{cao2025fault}. Common RNN variants include 
\replacetext{long short-term memory}{Long Short-Term Memory}
(LSTM) cells and 
\replacetext{gated recurrent units}{Gated Recurrent Units}
(GRU\newtext{s}). 
Faults in power systems can typically be inferred from a relatively limited time window of measurements.
Given the simpler architecture of GRUs, which are also computationally more efficient than LSTMs~\cite{cahuantzi2023comparison}, we will use GRU cells as part of our RGNN model to extract temporal dependencies.
To assess the contribution of the GNN part in our RGNN pipeline, we will benchmark them against pure RNN solutions, namely
\begin{enumerate*}[(i)]
    \item a \replacetext{separate}{shared} \emph{GRU} model \newtext{applied in parallel to all the PMUs individually}, to locally infer a failure (and then use majority voting to assess whether a fault occurs), and 
    \item an \emph{aggregated GRU} model, where we \replacetext{join the per-monitor GRUs}{apply a shared GRU model to each PMU node and then join their outputs} with a max pooling layer to output a single fault classification.
\end{enumerate*}

\subsection{Spatial Feature Extraction --  Graph Neural Networks}
\label{subsec:sfe_gnn}
A distribution network is naturally represented as a graph \( G = (V, E, A, H) \), where \( V \) is the set of \( N \) nodes (buses), i.e., \( |V| = N \), and \( E \) is the set of edges (branches) representing physical connections between nodes. The network structure is captured by the adjacency matrix \( A \in \mathbb{R}^{N \times N} \) (with $A_{uv} = 1$ if nodes $u$ and $v$ are connected by an edge).
In GNNs, each node \( v \in V \) is associated with a feature vector \( H_v \in \mathbb{R}^{F} \), forming the node feature matrix \( H \in \mathbb{R}^{N \times F} \). 

Graph learning aims to map node and edge attributes to target outputs to target outputs, 
modeled as:
\begin{equation}
    \hat{y} = f(G, \theta),
\end{equation}  
where \( \theta \) represents the model parameters, and \( \hat{y} \) is the predicted output.
GNNs 
iteratively aggregate 
information 
across nodes, in multiple layers: 
\begin{equation}
    H^{(k+1)} = f(H^{(k)}, A; \theta), \label{eq:gnn-general}
\end{equation}
where \( H^{(k)} \in \mathbb{R}^{N \times d_k} \) is the node representation matrix.
This is composed of row vectors $h^{(k)}_v$, representing each node $v \in V$ with a $d_k$-dimensional feature vector at layer $k$ (typically the same $\forall k$).
This iterative aggregation process can be viewed as a message passing framework, where each node $u$ updates its representation by aggregating the features from its neighbors $v \in \mathcal{N}(u)$.
GNN variants vary in how exactly this aggregation happens and what non-linear activation functions (which we will note as $\phi(\cdot)$) are used to obtain $f$ in \cref{eq:gnn-general}.
In our work, we consider four GNN variants:
\subsubsection{Graph Convolutional Network (GCN)} aggregates neighbor features using a normalized adjacency matrix $\hat{A}$~\cite{kipf2016semi},
updating the feature matrix $H^{(k+1)}$ as:
\begin{align}
    H^{(k+1)} &= \phi \left( \hat{A} H^{(k)} W \right), \quad W \in \mathbb{R}^{d_k \times d_k}
    \label{eq:eq6} 
    \\
    \text{with } \hat{A} &= D^{-1/2} A' D^{-1/2}, \quad \hat{A} \in \mathbb{R}^{N \times N}. \label{eq:eq5}
\end{align}
\noindent
Here $W$ is a learnable weight matrix, $A' = A + I$, $D$ is the degree matrix,
and \( d_k \) is the dimensionality of the node feature vector at layer \( k \). 
Note that this assumes a fixed topology, defined through $A$.

\subsubsection{GraphSAGE (Graph Sample and Aggregation)}~\cite{hamilton2017inductive}
updates a node’s representation using a learned aggregation function that 
does not explicitly rely on the graph structure (as opposed to GCNs, which need a fixed $A$).
This implies that a trained GraphSAGE can easily be used for a different graph topology at inference.
The updated feature vector for node $v$ at layer $k+1$ is given by:\footnote{The original GraphSAGE model alternatively proposes a sampled subset $\mathcal{N}'(v) \subset \mathcal{N}(v)$ rather than the full set of neighbors. Given the relatively small size of power distribution networks (as opposed to \eg social media graphs), we apply GraphSAGE in its full-batch setting without any subsampling.} 
\begin{equation}
    h_v^{(k+1)} = \phi \left( W  \cdot\, \text{concat}\left( h_v^{(k)}, \text{AGG}\left( \{ h_u^{(k)} | u \in \mathcal{N}(v) \} \right) \right) \right),
\end{equation}
where $W$ is a learnable weight matrix, $\text{concat}$ is the concatenation operation, and $\text{AGG}$ represents an aggregation function. 
Common options for $\text{AGG}$ include max pooling, average pooling, and more complex aggregators such as LSTMs.

\subsubsection{Improved Graph Attention Network (GATv2)} introduces an attention-based message passing mechanism that adaptively weighs the contributions of neighboring nodes~\cite{brody2022attentive}. The representation of a node $v$ at layer $k+1$ is computed as:
\begin{align}
    h_v^{(k+1)} &= \phi \left( \sum_{u \in \mathcal{N}(v)} \alpha_{vu} W h_u^{(k)} \right) \quad\text{with}\\
    \alpha_{vu} &= \frac{\exp \left( a^T \cdot \text{LeakyReLU}(W_1 h_v^{(k)} + W_2 h_u^{(k)}) \right)}{\sum_{j \in \mathcal{N}(v)} \exp \left( a^T \cdot \text{LeakyReLU}(W_1 h_v^{(k)} + W_2 h_j^{(k)}) \right)}, \label{eq:GATv2}
\end{align}
where $a$ is a learnable attention vector and $\text{LeakyReLU}$ denotes a nonlinear activation. This attention mechanism allows the model to focus on the most relevant neighbors based on the learned coefficients.
The earlier Graph Attention Network (GAT) proposed by Velickovic et al.~\cite{velickovic2018graph} computed attention as a linear combination of features ($W h_v^{(k)} || W h_u^{(k)}$) which applies the same linear transformation $W$ to both nodes before concatenation, and the nonlinearity is applied after attention is computed.
In contrast, improved GAT(GATv2) provides (i) more flexibility by using separate learnable weights ($W_1 h_v + W_2 h_u$) for each node, which accounts for asymmetric neighbor contributions, and (ii) greater expressiveness by applying nonlinearity before attention calculation.

\begin{figure*}[t]
    \centering
    \includegraphics[width=\textwidth]{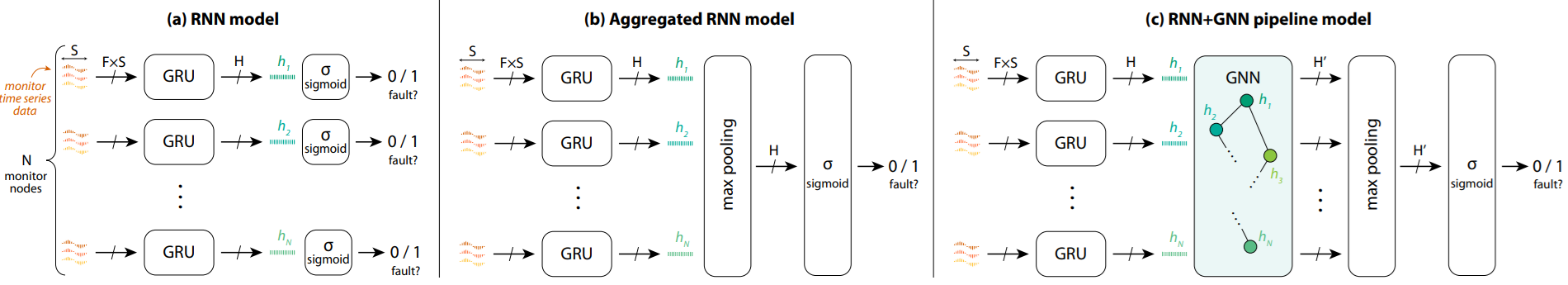}
    \caption{Illustration of the (a)~\replacetext{per-monitor}{shared} RNN \newtext{for local inference per PMU}, (b)~aggregated RNN model, and (c)~RNN+GNN pipeline architecture. The GNN layers consist\oldtext{s} of message-passing, dropout, and batch normalization, with each variant using its respective architecture (see \cref{subsec:sfe_gnn}).
    Further, $N$ represents the number of \replacetext{monitor}{PMU} nodes, $F$ is the number of features (\eg phase voltages, currents), $S$ is the time series sequence length, $H$ denotes the dimension of the GRU hidden state representation (and hence output), and $H'$ is the dimension of the GNN layer's output node representations.
 }
    \label{fig:rgnn_architecture}
\end{figure*}

\subsection{Proposed RGNN framework}

In \newtext{the} literature, both RNNs and GNNs have been adopted individually for fault diagnosis in power distribution systems, \eg respectively in~\cite{yu2019intelligent} and~\cite{chen2020fault}. 
However, this means the model either focuses on extracting temporal (RNN) or spatial (GNN) information, but not both.
Hence, combining RNNs and GNNs into an RGNN have been proposed 
to jointly model temporal and spatial dependencies.
In our work, we implement an RGNN pipeline where the input is first processed by a GRU cell to capture temporal dependencies, followed by a GNN model to extract spatial relationships.
We \replacetext{particularly will}{will particularly} explore various recent GNN models.
For that GNN part, we are the first to consider GraphSAGE and graph attention (GAT, GATv2) models --- which
incorporate
attention mechanisms and inductive learning capabilities, respectively --- for the considered power distribution fault diagnosis application.
\Cref{fig:rgnn_architecture} summarizes our proposed pipeline architecture.
For each of the $N$ \replacetext{monitor}{PMU} nodes, the input time series of $S$ timesteps (see details in \cref{sec:simulation-setup}), is processed by a GRU.
The node representations given by the GRU outputs are then processed by a GNN model (with the \replacetext{monitors}{PMUs} being the nodes and their connected edges defined based on the grid topology; details in \cref{sec:simulation-setup}).
The final GNN layer's node representations are then aggregated with a max pooling layer, and finally a sigmoid produces the single binary classification output.
 
\section{Experimental Setup}
\label{sec:exp}

\subsection{Fault Types}
\label{sec:fault-types}
Faults in power systems are typically classified into two types: open circuit and short circuit faults~\cite{gururajapathy2017fault}. Open circuit faults occur when conductors break, disrupting circuit continuity, while short circuit faults result from unintended contact between conductors or with the ground, causing abnormal current surges. Short circuit faults, especially due to their frequency and impact on system stability, are the focus of most detection studies. These faults can be further categorized into asymmetrical faults (line-to-ground (LG), line-to-line (LL), and double line-to-ground (LLG)), and symmetrical faults such as three-phase faults. LG faults are the most common, making up approximately  75\%–80\% of power system faults~\cite{glover2012power, mousa2019fault, gururajapathy2017fault}, and are the focus of our study.

\subsection{Simulation Setup and Data Collection}
\label{sec:simulation-setup}
\begin{figure}[t]
\centering
\includegraphics[width=\columnwidth]{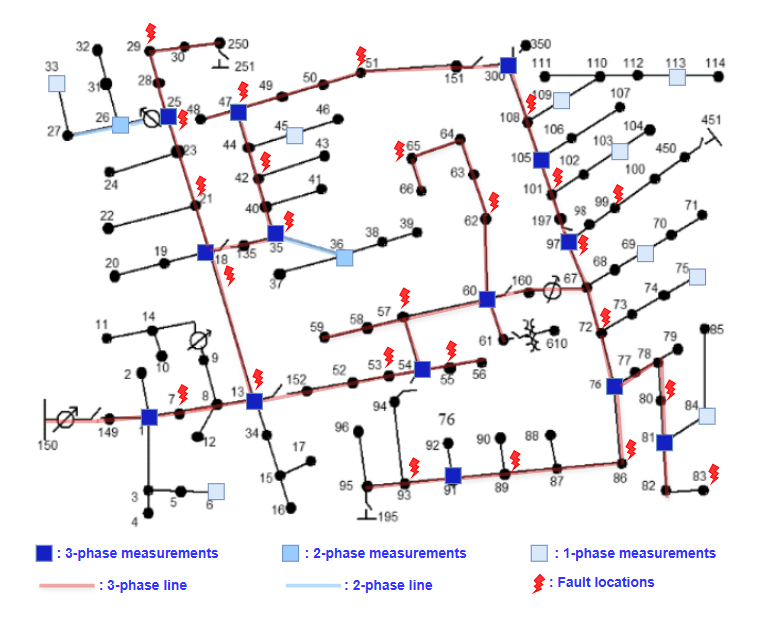} 
\caption{IEEE 123-node feeder with fault locations and voltage measurements.}
\label{fig:IEEE123}
\end{figure}

The proposed models are data-driven, and their performance depends on high-quality data.
Given the rarity of faults and the limited \replacetext{monitoring}{measurement} infrastructure to capture them, we simulate data using OpenDSS~\cite{epri_opendss} and the PyDSS interface~\cite{pydss}.
We conduct our experiments on the IEEE 123-bus feeder, a widely-used benchmark system in fault diagnosis studies\cite{nguyen2022one, nguyen2023spatial, chen2020fault}, operating at a nominal voltage of \SI{13.2}{\kilo\volt}  and a frequency of \SI{60}{\hertz}.
Fault durations typically range from \SI{20}{\milli\second} to \SI{50}{\milli\second} in distribution systems; for consistency, we consider a fault duration of \SI{20}{\milli\second}.
As indicated in \cref{fig:IEEE123}, faults are dynamically injected at $25$~locations, and measurements are gathered by $N\,=\,25$ \replacetext{monitors}{PMUs}.
We collect measurement data at \SI{1}{\milli\second} resolution, thus providing $20$ data points per fault for detailed fault profile analysis.

From the measurement data, we recorded \SI{60}{\milli\second} fault windows, starting \SI{40}{\milli\second} before the fault. 
As input to our classification models (GRU or RGNN), we feed a time series of $S = 20$ timesteps, by considering sliding windows of \SI{20}{\milli\second} from the collected samples. 
\newtext{From a total of 3 load scenarios, 25 fault locations, and  25 PMUs, 1,875 unique PMU data streams are collected. Each of these PMU's data} contributes 41 windows, 21 non-faulty and 20 faulty, resulting in a nearly balanced dataset mixture of faults and fault-free cases.
\Cref{table:dataset_info} summarizes our data generation configurations: in total, 76,875 sliding windows were collected, including 37,500 fault cases and 39,375 non-fault cases.
\oldtext{from three load scenarios and three fault resistance values ($\SI{0.1}{\ohm}$, $\SI{1}{\ohm}$, and $\SI{10}{\ohm}$).}
The full dataset was then randomly split into a \oldtext{set}{$\sim$68\% training, $\sim$17\% validation, and $\sim$15\% test set.}
 
\begin{table}[t]
\caption{Dataset Information} 
\centering
\resizebox{\columnwidth}{!}{ 
\begin{tabular}{lccc} 
\toprule
\toprule
\textbf{Elements} & \multicolumn{2}{l}{\textbf{Value}} & \textbf{Number} \\
\midrule
Fault type & \multicolumn{2}{l}{LG} & 1 \\
\midrule
Fault resistance (\si{\ohm}) & \multicolumn{2}{l}{0.1, 1.0, 10} & 3 \\
\midrule
Fault position (Buses) & \multicolumn{2}{p{6cm}}{7, 13, 18, 21, 25, 29, 35, 42, 47, 51, 53, 55, 57, 62, 65, 72, 80, 83, 86, 89, 93, 97, 99, 101, 108} & 25 \\
\midrule
\replacetext{Monitor}{PMU} location & \multicolumn{2}{p{6cm}}{1, 6, 13, 18, 25, 26, 33, 35, 36, 45, 47, 54, 60, 69, 75, 76, 81, 84, 91, 97, 103, 105, 109, 113, 300} & 25 \\
\midrule
Load Scenarios (p.u.) & \multicolumn{2}{l}{0.7, 1.0, 1.3}  & 3 \\
\bottomrule
\toprule
\multicolumn{4}{c}{\textbf{Dataset Distribution}} \\
\midrule
\textbf{Category} & \textbf{Train Set} & \textbf{Validation Set} & \textbf{Test Set} \\
\midrule
Fault Cases & \newtext{25.610} & \newtext{6,402} & \newtext{5,488} \\
\midrule
\replacetext{Non-Fault}{No Fault} Cases & \newtext{26,890} & \newtext{6,723} & \newtext{5,762} \\
\midrule
Total Cases & \newtext{52,500} & \newtext{13,125} & \newtext{11,250} \\
\bottomrule
\end{tabular}
} 
\label{table:dataset_info}
\end{table}

For evaluating model robustness to changing \replacetext{monitoring}{PMU} configurations, different \replacetext{monitoring}{PMU} configurations are subsampled from the entire 25-\replacetext{monitor}{PMU} setup (see \cref{table:dataset_subsets}).
We note that the configurations with a lower number of PMUs are subsets of those with a higher number.
Those with fewer PMUs include mostly higher-degree nodes, which should be more informative than peripheral ones, thus reflecting real-world PMU settings.
Modern distribution grids evolve due to DER integration (\eg solar panels, wind turbines, residential battery storage) and electrification (\eg EV chargers).
Since DERs are typically connected at leaf nodes, new measurement devices are often introduced at these points, while existing peripheral measurement devices may be retired or repurposed. Our approach reflects these dynamic updates in measurement infrastructure.

\begin{table}[t]
\caption{
List of PMUs used in each configuration. Nodes dropped from one subset to the next are typeset in \textcolor{red}{red}.}
\centering
\resizebox{\columnwidth}{!}{
\begin{tabular}{>{\centering\arraybackslash}m{1.5cm} >{\centering\arraybackslash}m{6cm}} 
\toprule
\toprule
\textbf{\# PMUs} & \textbf{Bus locations (see \cref{fig:IEEE123})} \\
\midrule
25 & 1, 6, 13, 18, 25, 26, 33, 35, 36, 45, 47, 54, 60, 69, 75, 76, 81, 84, 91, 97, 103, 105, 109, 113, 300  \\
\midrule
19 & 1, \textcolor{red}{\st{6}}, 13, 18, 25, 26, 33, 35, \textcolor{red}{\st{36}}, 45, 47, 54, 60, 69, 75, \textcolor{red}{\st{76}}, 81, \textcolor{red}{\st{84}}, 91, 97, \textcolor{red}{\st{103}}, 105, 109, \textcolor{red}{\st{113}}, 300  \\
\midrule
15 & 1, \st{6}, 13, 18, \textcolor{red}{\st{25}}, 26, \textcolor{red}{\st{33}}, 35, \st{36}, \textcolor{red}{\st{45}}, 47, 54, 60, 69, 75, \st{76}, 81, \st{84}, \textcolor{red}{\st{91}}, 97, \st{103}, 105, 109, \st{113}, 300  \\
\midrule
11 & 1, \st{6}, 13, \textcolor{red}{\st{18}}, \st{25}, 26, \st{33}, 35, \st{36}, \st{45}, 47, \textcolor{red}{\st{54}}, \textcolor{red}{\st{60}}, \textcolor{red}{\st{69}}, 75, \st{76}, 81, \st{84}, \st{91}, 97, \st{103}, 105, 109, \st{113}, 300  \\
\midrule
7 &  \textcolor{red}{\st{1}}, \st{6}, 13, \st{18}, \st{25}, \textcolor{red}{\st{26}}, \st{33}, 35, \st{36}, \st{45}, 47, \st{54}, \st{60}, \st{69}, \textcolor{red}{\st{75}}, \st{76}, 81, \st{84}, \st{91}, \textcolor{red}{\st{97}}, \st{103}, 105, 109, \st{113}, 300  \\
\bottomrule
\end{tabular}
}
\label{table:dataset_subsets}
\end{table}


\subsection{Model Training and Evaluation}
\label{sec:model-training}
We used Pytorch with the PyG library~\cite{fey2019fast} for GNNs.
We used three-phase voltages, currents, and their respective phase angles as features and applied Z-score feature normalization  (resulting in a mean of $0$ and a standard deviation of $1$).
Our GRU and GNN layers have a hidden state size of $128$, and we adopt binary cross-entropy with logits \oldtext{loss is used} as loss function for our fault classification objective.
To alleviate overfitting, we apply dropout and batch normalization, and attention dropout for the GAT models. 
All models are trained for $35$ epochs using the AdamW optimizer~\cite{llugsi2021comparison}. 

For the GNN part, we only consider the $N$ \replacetext{monitored}{measured} buses as graph nodes, ensuring that the graph is tailored for fault detection tasks and aligns with real-world \replacetext{monitoring}{measurement} constraints.
Other works~\cite{sun2021distribution, chen2020fault} include \replacetext{unmonitored}{unmeasured} buses with zero feature values, which may be beneficial for fault localization but could lead to noise propagation and over-smoothing issues~\cite{rusch2023survey}.
Excluding \replacetext{unmonitored}{unmeasured} buses also limits the graph size and computational complexity. 
In the GraphSAGE-based models (RGSAGE), given the small graph size ($N \le 25$)\newtext{,} we use the full-batch setting and adopt either max or mean pooling as \newtext{the} aggregation function (as a lightweight yet effective alternative to more complex methods such as LSTMs~\cite{hamilton2017inductive}).

\begin{figure*}[hbtp]
    \centering
    \includegraphics[width=.85\textwidth]
    {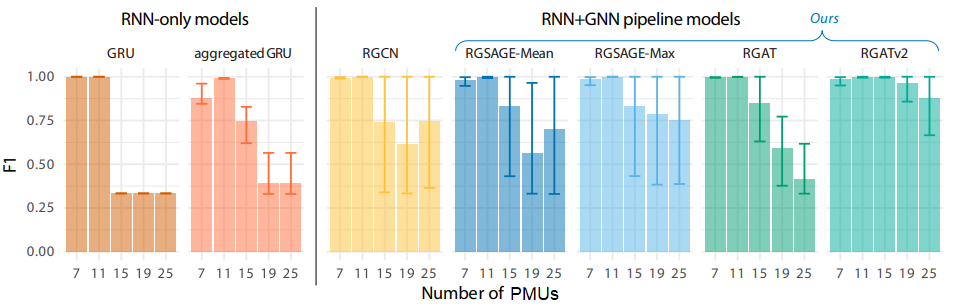}
    \caption{F1 Scores for Fault Event Detection using RNN-only models (GRU-based; left) and GNN-based models (right). All models are trained on the 11-\replacetext{monitor}{PMU} setup and evaluated across all configurations (7--25 \replacetext{monitors}{PMUs}; see \cref{table:dataset_subsets} for details). Error bars show the 90\% confidence intervals across 5~models trained with different random seeds.}
    \label{fig:F1-scores}
\end{figure*}

\section{Results and Discussion}
\label{sec:res}
We now discuss the quantitative performance analysis of the various RNN and RNN+GNN models, as performed on the IEEE 123-bus topology. 
\Cref{fig:F1-scores} shows the various architectures' fault detection performance (\ie binary classification fault \vs no fault), for models trained on the 11-\replacetext{monitor}{PMU} setup.

Let's first focus on comparing the various models' fault detection performance, when testing them in similar conditions as they were trained for.
Looking at \cref{fig:F1-scores}, the test results for 11-\replacetext{monitor}{PMU} setup (\ie what the model was trained for) show that all models, including GRUs, perform well on this relatively simple fault detection task ($\text{F1} \simeq 1$).
This suggests that GNN-based models' conceptual strengths do not shine for this relatively simple classification problem. 
Yet, we hypothesize that GNNs may offer benefits over when we consider more advanced diagnosis tasks (\eg fault localization, see also \cref{sec:conclusion}).

Now, more interestingly, let's focus on the models' \emph{generalization} capabilities and consider test results for a different number of \replacetext{monitors}{PMUs} than the trainin\newtext{g} set ($N \neq 11$).
First, for the case of \emph{fewer} \replacetext{monitors}{PMUs} ($N =7$), we note that
all methods mostly maintain high performance ($\text{F1}$ still close to $1$) --- which is  intuitive, as the $N=7$ \replacetext{monitors}{PMUs} used at inference, are included in those seen during training. 
Second, switching to cases with \emph{more} PMUs, \ie new PMU nodes are included at test time ($N > 11$), 
we observe that the RRN-only models (GRU, aggregated GRUs) largely fail with a significant performance drop ($\text{F1} \le 0.4$ on average).
In contrast, the RGNN pipeline models all perform quite stronger. 
This better generalization capacity is what we intuitively expect from their ability to process topological relationships (\ie structure and connectivity of the graph) to aggregate the distributed inputs into a global view.

Still, not all GNN architectures perform equally well.
\begin{enumerate*}[(i)]
    \item The \emph{RGCN} variant performs the most poorly even for a small number of added \replacetext{monitors}{PMUs} (\ie $N=15$). \replacetext{Possibly, this is}{This is possibly} due to GCN relying on a fixed topology (through the fixed adjacency matrix $A$).
    \item Our proposed \emph{RSAGE} approach performs slightly better (than RGCN), particularly when using max pooling.\footnote{That max pooling works better for detecting faults --- of which the effects may differ quite a bit from one monitor to the next --- intuitively was to be expected.} Still, there is \oldtext{still}a significant performance drop (down to $\text{F1} \simeq .75$).
    \item Our second proposal, of using \emph{GATv2}, is the only one that maintains performance, even for test results at $N=25$, where the majority of \replacetext{monitors}{the PMUs} \replacetext{was}{were} unseen during training. The asymmetric processing across GNN edges (\cf \cref{eq:GATv2}) --- which distinguishes it from the original GAT --- seems crucial to effectively cater for different node behavior. Indeed, the added \replacetext{monitors}{PMUs} for $N=25$ are mostly leaf nodes (see \cref{fig:IEEE123}), less centrally located than the initial case of $N=11$ \replacetext{monitors}{PMUs}  seen during training.
\end{enumerate*}

In summary, our results show that GNN-based models are more robust compared to pure RNN-based models for distribution grid fault detection. In particular, in terms of generalization to unseen \replacetext{monitor}{PMU} configurations, our proposed RGATv2 outperforms both traditional RNN-based models and \newtext{the} state-of-the-art RGCN variant.

\section{Conclusion and Future Work}
\label{sec:conclusion}
The aim of this paper’s pilot study was to comprehensively and systematically \begin{enumerate*}[(i)]
    \item\label{it:analyze-comprehensively}present and compare the various GNN-based models for distribution grid fault detection, and particularly \item\label{it:analyze-generalization}analyze the generalization capacity of those models when deploying them in configurations different from those seen at model training time.
\end{enumerate*}
For \ref{it:analyze-comprehensively}, we consider the IEEE 123-bus system, and train models for the simple binary classification case (fault \vs no fault).
In term\newtext{s} of models, we compare advanced RNN+GNN models (RGNN in short) as well as pure RNN-based baselines. 
In particular, we are the first to propose GraphSAGE and \newtext{G}raph \newtext{A}ttention (GAT/GATv2) models for this application.
For \ref{it:analyze-generalization}, we consider models trained on a $N=11$ \replacetext{monitor}{PMU} setup and then test for $N \neq 11$.



Our pilot study indicates \replacetext{first of all that}{that, first of all,} for fixed deployments, where models are trained and tested under the same conditions (with $N=11$ in our experiments), all considered models (including pure RNN models) are adequate.
Still, we hypothesize this may no longer be the case when moving to more challenging diagnosis applications.
Therefore, our future work aims to study fault type classification and particularly fault localization.
Nevertheless, the current pilot experiments studying generalization, considering a varying number of \replacetext{monitor}{PMU} nodes, already highlighted the potential benefits of GNN-based approaches.
Specifically, our newly proposed adoption of the RGATv2 model for fault detection demonstrated superior performance. 
While pure RNN-based solutions largely fail (F1 below 50\%)  to perform when adding new \replacetext{monitor}{PMU} nodes, even the previously proposed RGCN model as well as our other proposal, RGSAGE, suffers drops in F1-score of $\sim$25\%.

As indicated above, we plan to extend this pilot study to more detailed fault diagnosis tasks, especially fault location.
Moreover, we will extend the generalization scenarios to a broader scope of topological variations (\eg switching operations, network reconfigurations), in light of recent evolutions \replacetext{in}{of} today's power grid.
For instance, as opposed to the more conventional fault scenarios studied above, distributed energy resources (DERs) --- typically renewable sources such as photovoltaics --- can introduce larger variability in fault characteristics, leading to more complex fault patterns. 

\section*{Acknowledgment}
Research reported in this publication was supported by VITO grant number VITO\_UGENT\_PhD\_2301.
This research was also partially funded by the Flemish Government (AI Research Program). 

\bibliographystyle{ieeetr} 
\bibliography{references}

\newpage 
\onecolumn

\end{document}